\documentclass[sigconf]{acmart}

\AtBeginDocument{%
  \providecommand\BibTeX{{%
    \normalfont B\kern-0.5em{\scshape i\kern-0.25em b}\kern-0.8em\TeX}}}

\copyrightyear{copyrightyear}
\acmYear{acmyear}
\setcopyright{rightsretained}
\acmConference{acmconference}
\acmBooktitle{acmbooktitle}
\acmDOI{acmdoi}
\acmISBN{acmisbn}

\begin{document}

\title{MobileCodec: Neural Inter-frame Video Compression \\ on Mobile Devices}

\author{Hoang Le, Liang Zhang, Amir Said, Guillaume Sautiere, \\ Yang Yang$^\ddagger$, Pranav Shrestha, Fei Yin$^\ddagger$, Reza Pourreza, Auke Wiggers}
\affiliation{%
  \institution{Qualcomm AI Research}
  \streetaddress{*}
  \city{}
  \country{}}

\renewcommand{\shortauthors}{Le et al.}

\begin{abstract}
Realizing the potential of neural video codecs on mobile devices is a big technological challenge due to the computational complexity of deep networks and the power-constrained mobile hardware. 
We demonstrate practical feasibility by leveraging Qualcomm's technology and innovation, bridging the gap from neural network-based codec simulations running on wall-powered workstations, to real-time operation on a mobile device powered by Snapdragon\textsuperscript{\tiny\textregistered} technology. 
We show the first-ever inter-frame neural video decoder running on a commercial mobile phone, decoding high-definition videos in real-time while maintaining a low bitrate and high visual quality.
\end{abstract}


\maketitle
\let\thefootnote\relax\footnote{
Snapdragon and Qualcomm AI Engine are products of Qualcomm Technologies, Inc. and/or its subsidiaries.
Qualcomm AI Research is an initiative of Qualcomm Technologies, Inc.}
\let\thefootnote\relax\footnote{
$\ddagger$: This work was completed during employment at Qualcomm Technologies, Inc. \vspace{-10mm}}

\section{Introduction}
Video compression technologies have been actively researched and engineered over the past decades to obtain broad video adoption across a wide range of devices, distribution media, and services.
As neural network (NN) technologies have been revolutionizing the world, NN-based video coding methods, especially deep generative model-based approaches, have a strong potential to further extend the capabilities and efficiency of video compression. 
Instead of manually designing a sophisticated pipeline with multiple stages and optimizing each stage separately, a NN-based codec can learn to directly extract and code a low-dimensional feature representation of the input data in an end-to-end process, offering a better rate-distortion tradeoff. 
Moreover, it is easy to upgrade and deploy new NN-based codecs, as their models are trained in a relatively short amount of time in comparison to the development of conventional video codecs. 
Equally important is that neural codecs can leverage general-purpose AI hardware accelerators, which are already ubiquitous in many different computing platforms. 

Recent advances in deep learning have revolutionized the way lossy data compression algorithms are designed. 
Specifically, we have seen remarkable success in image and video compression, achieved by employing a convolutional autoencoder as a flexible learned non-linear transform~\cite{balle2017iclr, balle2020nonlinear}, deep generative models as powerful entropy models~\cite{balle2018variational, NEURIPS2018_53edebc5}, and various neural motion estimation and compensation algorithms for high-quality frame interpolation and extrapolation~\cite{NEURIPS2019_bbf94b34,NEURIPS2020_9a118833}. 
After a few years of development, neural video codecs~\cite{lu2019dvc, Agustsson2020-ro,b_epic, hu2021fvc, rippel2021elfvc, hu2022coarse} are now on par with or even outperform handcrafted solutions such as HEVC~\cite{6316136}, and have become a vibrant research area with consistent year-over-year improvements in rate-distortion performance.

Besides competitive rate-distortion performance, learnable codecs also open new opportunities such as easy adaptation to different data distributions~\cite{van2021overfitting,strumpler2021implicit}. 
Their flexibility allows optimization of subjective quality of reconstructed video as well, for example using learned region-of-interest coding to allocate more bits to semantically interesting regions \cite{perugachi2022region}, or the use of differentiable perceptual metrics ~\cite{NEURIPS2020_8a50bae2, mentzer2021towards, yang2021perceptual}. 

However, most of the studies use wall-powered high-end GPUs with floating-point computation, and the used neural network models are often not optimized for fast inference.
As a result, most neural video codecs are unable to run in real-time on power-constrained devices such as mobile phones.
This is a hurdle for adoption of neural codecs, as a large part of video communication today takes place on mobile.

In this paper, we therefore demonstrate that neural video decoding on a mobile phone is feasible.
Leveraging the Qualcomm\textsuperscript{\tiny\textregistered} AI Engine, we achieve the first real-time neural inter-frame decoder running on a commercial mobile device. 
The main contributions of this work are:
\begin{itemize}
    \item A demonstration of the first real-time neural inter-frame video decoder on a mobile device.
    \item An efficient neural inter-frame codec architecture, specifically designed for deployment on a mobile platform.
    \item A parallel entropy coding algorithm tailored for neural entropy models.
\end{itemize}

\begin{figure*}
\includegraphics[width=5.6in]{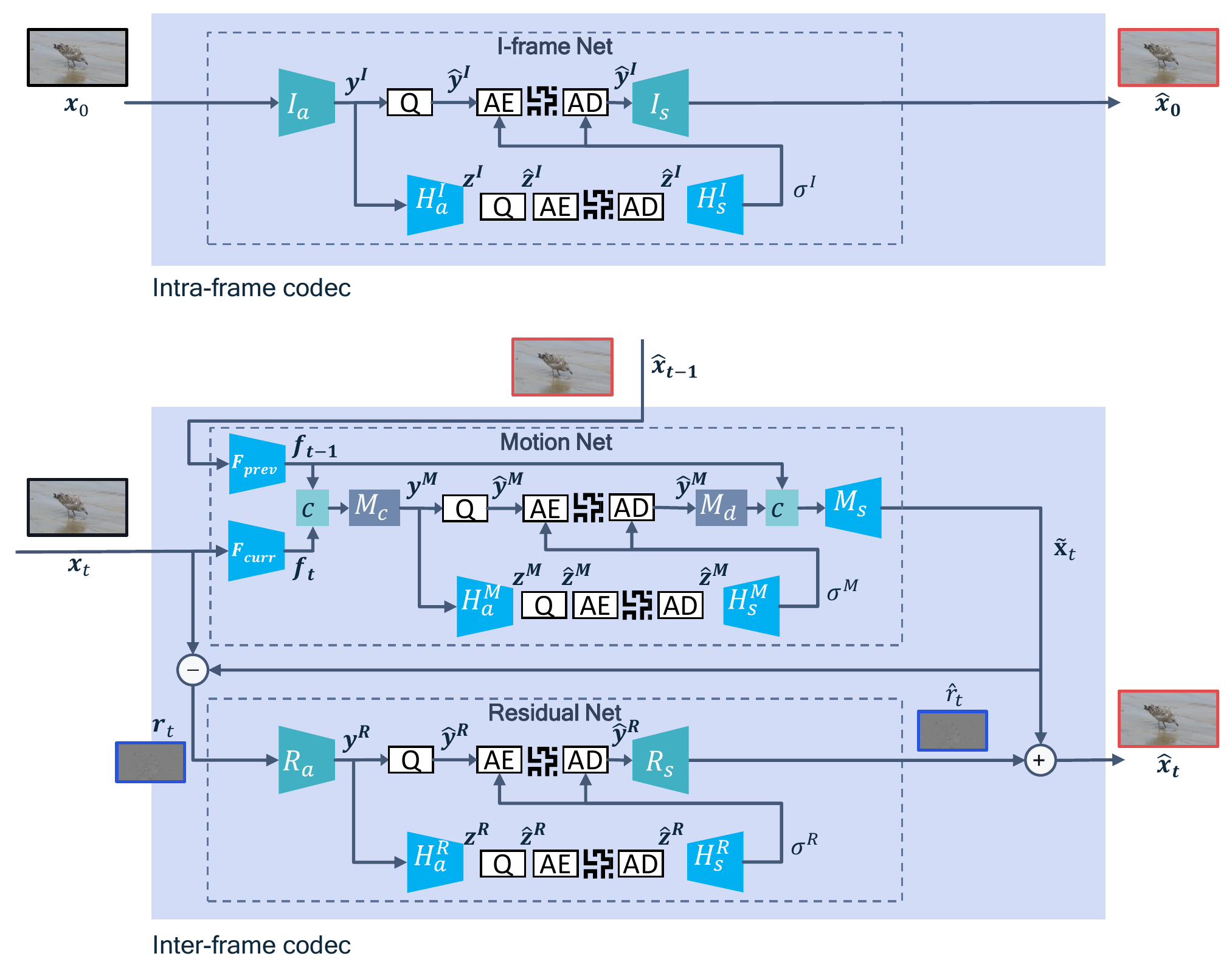}
\centering
\vspace{-0.3cm}
\caption{MobileCodec network architecture ~\label{fig:archi}}
\end{figure*}

\section{Related Work}

\noindent\paragraph{Neural image compression}
There has been significant progress in the development of neural image compression solutions in recent years. 
While earlier works proposed different types of generative models ~\cite{todericiVariableRateImage2015, todericiFullResolutionImage2017, rippel2017real,mentzerConditionalProbabilityModels2018}, many recent works are based on the hierarchical
hyperprior architecture~\cite{balleDensity2015,balleEndtoendOptimizationNonlinear2016,balle2018variational,lu2021progressive,zhu2021transformer}. 
Its variants equipped with autoregressive models~\cite{NEURIPS2018_53edebc5,minnen2020channelwise} and attention mechanisms~\cite{liuNLAICImageCompression2019} are currently the most widely adopted architectures.

\noindent\paragraph{Neural video compression} 
Neural video compression has seen advances for both low-latency video~\cite{liu2020learned,lu2019dvc,DVC_TPAMI} and video streaming applications~\cite{wuVideoCompressionImage2018,Cheng_2019_CVPR,Habibian_2019_ICCV,Djelouah_2019_ICCV,b_epic}. 
The low-latency solutions often include unidirectional motion estimation/compensation followed by a residual correction~\cite{chen2020learning,Agustsson2020-ro}. 
Recent works have introduced recurrency~\cite{Golinski_2020_ACCV,rippelLearnedVideoCompression2018} or longer range dependencies~\cite{mlvc, rippel2021elfvc, hu2021fvc, hu2022coarse} to improve performance. 
In the settings where it is acceptable to incur latency by using reference frames from future timepoints, the existing solutions include bidirectional motion estimation and compensation~\cite{Djelouah_2019_ICCV, b_epic}, frame interpolation~\cite{wuVideoCompressionImage2018,Cheng_2019_CVPR} followed by residual correction, or implicit multi-frame coding approaches based on 3D convolutions~\cite{Habibian_2019_ICCV}. 

\begin{figure*}[t!]
\includegraphics[height=1.60in]{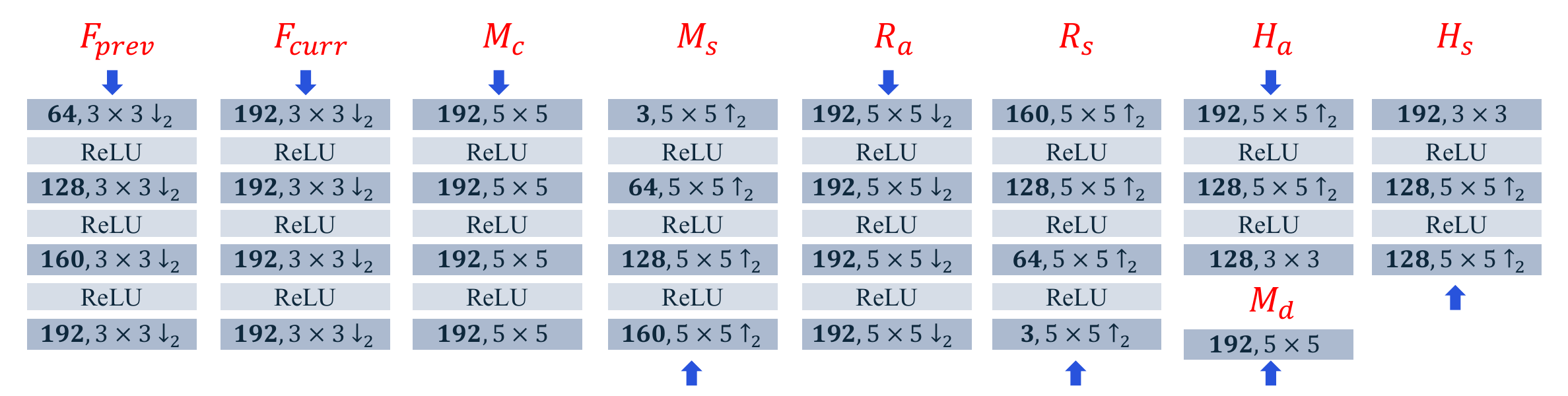}
\centering
\vspace{-0.25cm}
\caption{Asymmetric efficient network architecture~\label{fig:archi_details}. A network used in the receiver uses smaller number of filters than a network used in the transmitter. We use fewer filters at high resolution to reduce computation as well.
}
\end{figure*}

\noindent\paragraph{Efficient neural codecs}
Despite the progress in developing neural image and video compression solutions, literature that explicitly aims to improve their efficiency and practical usefulness is limited.
Efficient architectures are explored in~\cite{liu_2020_eccv,rippel2017real,rippel2021elfvc,Johnston2019-re}.
Overfitting to a video or segments of a video is another means to reduce model complexity while improving rate-distortion performance~\cite{Rozendaal_2021_iclr, strumpler2021implicit, zhang2021implicit}.
Yet, in their current form, these methods are unable to run in real-time on resource-constrained mobile devices.
To the best of our knowledge, these exists only a single public neural video codec deployed to a mobile device~\cite{intra_demo} that runs in all-intra mode, i.e., all frames are coded as images.

\section{Approach}

We enable real-time neural inter-frame decoding using three main steps. 
First, we design a mobile-friendly neural network architecture, \textbf{MobileCodec}, optimized for efficient inference on a mobile device.
Second, we use quantization-aware training~\cite{Nagel2021-eo} to quantize network parameters and activations to fixed point, retaining the networks original floating-point performance. 
Finally, we develop a parallel entropy coding algorithm that leverages the parallel inference property of neural network-based codecs. 
We describe these steps in more details in the following sections. 

\subsection{Mobile friendly network architecture\label{sec:approach:archi}}
To optimize MobileCodec for performance on a mobile device, we use a network consisting of mostly convolutional layers and ReLU activations, as both operations are extremely efficient. 
Figure~\ref{fig:archi} shows the overall design of our network architecture. 
The system consists of two codecs: an intra-frame codec that compresses each video frame independently (as an image), and an inter-frame codec that compresses a frame conditioned on a previously decoded frame. 
The intra-frame codec is used to compress the first frame of each video segment, and the inter-frame codec is used to compress the rest of the frames of that segment.
Such a segment is typically referred to as a Group of Pictures (GoP).

\subsection{Intra-frame MobileCodec}
The design of our Intra-frame codec is derived from the image compression method by Balle et al.~\cite{balle2018variational}. 
Please refer to the original paper for a detailed description of the method. 
Summarizing, as shown in Figure~\ref{fig:archi}, the encoder (or analysis transform) $I_a$ transforms each frame $x_0$ into a latent variable $y^I = I_a(x_0)$. 
This latent $y^I$ is then quantized by a rounding operator $Q$, and entropy coded by an arithmetic encoder ($AE$).
The resulting bitstream is sent to the receiver, where it will be decoded by an arithmetic decoder ($AD$) and reconstructed back to obtain a reconstruction $\hat{x}_0$ via the synthesis decoder $I_s$: $\hat{x}_0 = I_s(y^I)$. 
Following~\cite{balle2018variational}, we also use a hypercodec network to model the probability distribution of the encoded latents $y^I$. 
Specifically, the hyper-encoder $H^I_a$ transforms the latent $y^I$ into a latent variable $z^I = H^I_a(y^I)$, referred to as a hyper-latent. 
Hyper-latent $z^I$ is also quantized, entropy coded, and sent to the receiver where it will be decoded and fed into a hyper-decoder network $H_s^I$ to predict the scale $\sigma^I$ of the distribution over latents $\hat{y}^I$: $\sigma^I = H^I_s(\hat{z}^I)$. 
The predicted scale $\sigma^I$ characterizes a distribution $p(\hat{y}^I)$, which is used for arithmetic encoding $AC$ only in the transmitter side, and for arithmetic decoding $AD$ in both transmitter and receiver side. 
Note that, in contrast to ~\cite{balle2018variational}, we as use ReLU as the nonlinearity after each layer of the network instead of a GDN block. 
We empirically observed that ReLU non-linearities are more quantization friendly than GDN blocks, and that they can be used without a drop in rate-distortion performance.

\subsubsection{Inter-frame MobileCodec}
For our inter-frame codec, we follow the recent design of neural inter-frame codec methods~\cite{Agustsson2020-ro, lu2019dvc} and include two consecutive sub-networks: a motion network and a residual network. 
First, the motion network takes a previously decoded frame $\hat{x}_{t-1}$ and the current frame $x_t$, and encodes the motion between them.
The decoded motion is used to obtain an initial reconstruction $\tilde{x}_t$ at the receiver end. 
The residual network encodes the difference $r_t$ between the reconstructed frame $\tilde{x}_t$ and the current frame $x_t$. 
This reconstructed residual $\hat{r}_t$ is then used to refine the reconstructed frame $\tilde{x}_t$ to obtain the final reconstructed frame $\hat{x}_t$.

\subsubsection{Flow-Agnostic Motion Compensation}
Motion compensation is an important component that contributes significantly in saving bitrate for an inter-frame codec.
Inter-frame codecs typically exploit redundancies between consecutive frames by transmitting the motion between them, as described earlier.
Most existing neural video coding approaches implement motion compensation either by pixel warping~\cite{Agustsson2020-ro, lu2019dvc} or feature-space warping~\cite{hu2021fvc}. 
These methods require large memory buffers to enable memory access to the arbitrary location of a video frame or the corresponding features.
As a result, these operations may not be efficient on a mobile device.
Moreover, these operations do not scale well with image resolution, as they allow arbitrary warping of pixels from anywhere in the frame. 
In contrast, convolutional operations operate on each image block and are highly efficient on mobile devices. 
For this reason, we design a flow-agnostic motion compensation network that uses only convolutional operations, thereby avoiding less efficient warping operations.
As illustrated in Figure~\ref{fig:archi}, our motion compensation method includes two feature extractors $F_{prev}$ and $F_{curr}$ to extract representative features $f_{t-1}$ and $f_t$ of the previously decoded frame $\hat{x}_{t-1}$ and the current frame $x_t$ respectively:
\begin{equation}
    f_{t-1} = F_{prev}(\hat{x}_{t-1}) \\
\end{equation}
\begin{equation}
    f_{t} = F_{curr}(x_t)
\end{equation}

The extracted features $f_{t-1}$ and $f_{t}$ are concatenated as input into the feature correlation module $M_c$ to extract the motion features between the two frames:
\begin{equation}
    y^M = M_c(concat(f_{t-1},f_{t}))
\end{equation}

These motion features are quantized via the quantizer $Q$, then coded via arithmetic coding $AE$, and sent to the receiver. 

At the receiver end, the decoded motion features $\hat{y}^M$ are concatenated with the feature $f_{t-1}$ extracted from the previous frame $\hat{x}_{t-1}$, which is available for both transmitter and receiver. These features are then fed into an image synthesis network $M_s$ to reconstruct the current frame $\tilde{x}_t$.

\begin{equation}
    \tilde{x}_t = M_s(concat(f_{t-1},\hat{y}^M))
\end{equation}

Next, the residual error $r_t=x_t-\tilde{x}_t$ of the reconstructed image $\tilde{x}_t$ is fed into a residual analysis encoder $R_a$ to extract the latent feature $y^R$.
\begin{equation}
    y^R = R_a(r_t)
\end{equation}

This latent feature $y^R$ is then also quantized via the quantizer $Q$ and entropy coded via $AE$, and sent to the receiver. On the receiver side, the decoded latent $\hat{y}^R$ is input into the residual decoder $R_s$ to reconstruct the residual $\hat{r}_t$. 
\begin{equation}
    \hat{r}_t = R_s(\hat{y}^R)
\end{equation}
The final reconstructed frame is generated by adding the reconstructed residual $\hat{r}_t$ to the motion compensated image $\tilde{x}_t$:
\begin{equation}
    \hat{x}_t = \tilde{x}_t + \hat{r}_t
\end{equation}

We follow \cite{balle2018variational} and use a hyperprior network to predict the scale $\sigma^M$ and $\sigma^R$ for each latent feature in $y^M$ and $y^R$ to be entropy coded respectively. 
Specifically, the hyperprior analysis encoder $H_a^M$ transforms the motion latent $y^M$ into motion hyperlatent $z^M$
\begin{equation}
    z^M = H^M_a(y^M)
\end{equation}
This motion hyperlatent $z^M$ is quantized, then entropy coded and sent to the receiver. At the receiver end, the decoded latent $\hat{z}^M$ is input into a hyperdecoder network $H_s^M$ to predict the scale $\sigma_M$ which is then used for entropy encoding and decoding for the latent $\hat{y}^M$. The same process is also employed for the entropy coding of the residual latent $\hat{y}^R$. 

\subsubsection{Asymmetric neural network architecture}
We use an asymmetric encoder-decoder architecture in order to enable real-time decoding.
This design is consistent with a recent video coding approach proposed in~\cite{rippel2021elfvc}, which showed that reducing decoder compute is possible without a reduction in rate-distortion performance. 
Figure~\ref{fig:archi_details} shows the details of our asymmetric architecture, and lists the number of channels and kernel size for each layer.

\subsection{Channel-wise Quantization Aware Training\label{sec:approach:qat}}
ALthough using 32 bit floating-point operations is a default operating mode on wall-powered computing devices like workstations or training servers, it is often more efficient to run in lower bitwidths or fixed point, especially on power-constrained devices. 
To increase runtime and improve power efficiency, we need to quantize the network parameters and activations to low-precision integers. 
However, naive post-training quantization of neural network to 8 bit integers would significantly reduce the rate-distortion performance. 
To alleviate this issue, we finetune our MobileCodec using the quantization-aware training (QAT) pipeline described in ~\citet{Nagel2021-eo}.
This allows learning the quantization binwidth for each layer weight and activations by introducing simulated straight-through gradient estimation. 

We quantize the entire model to 8-bit fixed point.
For the convolutional weights, we use separate quantization parameters for each output channel (also known as per-channel quantization) \cite{nagel2019datafree}, as we observe the range of both weights and activations varies widely between channels. 
This observation is in line with recent work~\cite{minnen2020channelwise} which shows that different channels of the latent impact the compression quality differently.

In addition, we pay particular attention to the variables that are important for entropy coding. 
The scale $\sigma$ in the hyperprior must cover several orders of magnitude, and would require very high precision if directly quantized. 
A better approach is to have the hyperprior networks learn an entropy coding parametrization suitable for low-precision representations. 
This enhanced entropy coding approach is described in more detail in ~\cite{Said:22:ppq}. 
For this video coding demo, a simplified version of that approach was used, where the hyperencoder and hyperdecoder learn a logarithmic scale (i.e., they learn $\ln(\sigma)$ instead of $\sigma$), which is then quantized to a single 8-bit integer in order to match pre-computed lookup tables (see Section \ref{sec:approach:pec}). 
The value of this 8-bit integer $n$ that is fed to the entropy encoder and decoder is computed using:
\begin{equation}
    n = \max\left[0, \min\left[255, \lfloor\gamma \ln({\sigma})\rfloor + \theta\right]\right] \label{eq:ECQuant}
\end{equation}
We set $\sigma=32$ and $\theta=70$ to cover the range of $\sigma$ needed for the demo. Reference~\cite{Said:22:ppq} presents more examples of alternative parametrizations.

\begin{figure}[t!]
\centering
\includegraphics[width=3.27in]{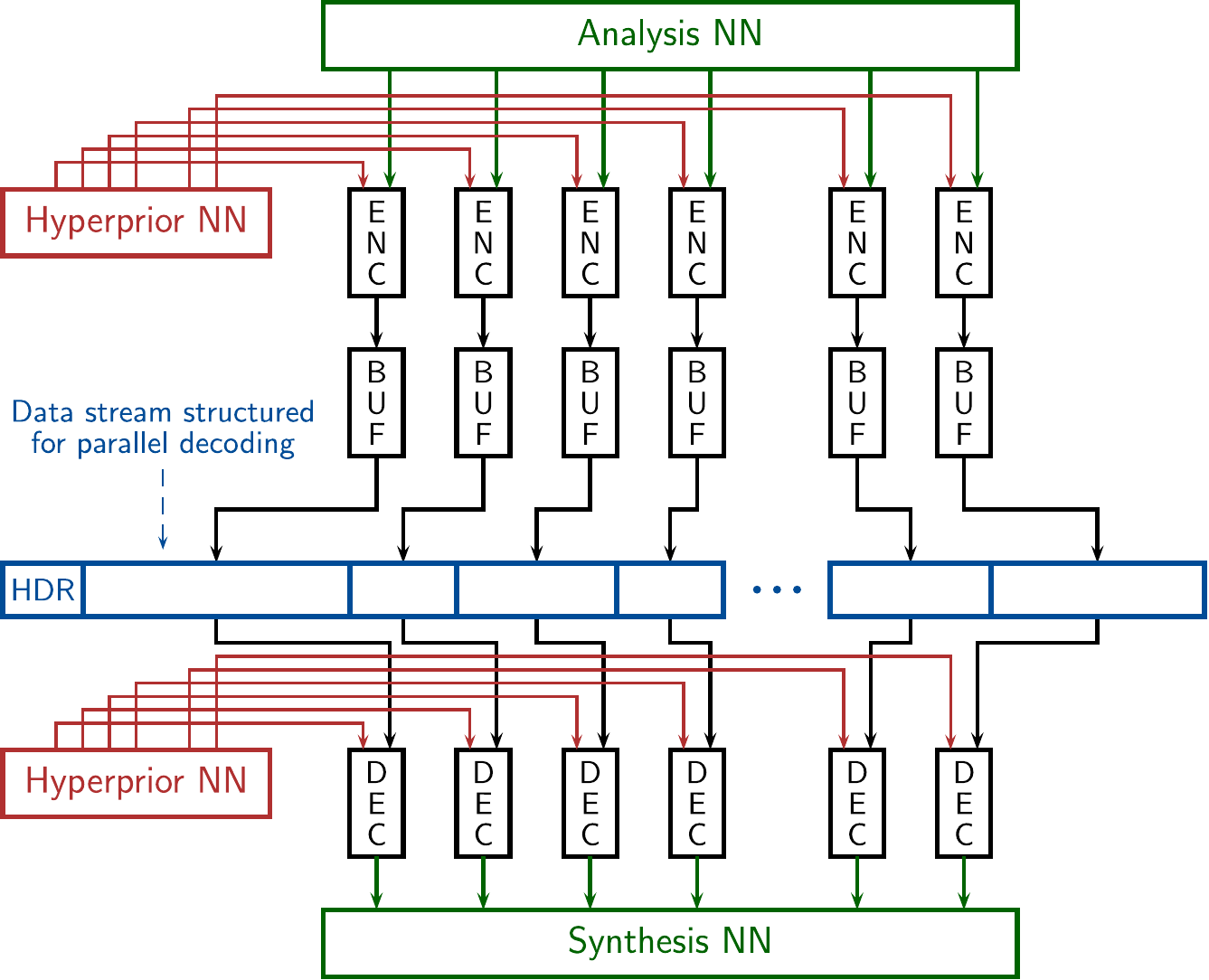}
\caption{Parallel entropy coding for our neural video codec~\label{fig:pec} (excluding data stream with hyperprior latents).}
\end{figure}

\subsection{Parallel Entropy Coding\label{sec:approach:pec}}

Entropy coding is employed to losslessly compress the quantized latents and hyperlatents, and further reduce the bit rate. 
This is the stage where the bitstream representing the video is created (at the transmitter) or parsed (at the receiver). 
For high resolution and quality levels, entropy coding must support very high throughputs, which can be difficult and computationally expensive without parallelization, even with custom hardware. 

As the probability models in our MobileCodec are designed to be (conditionally) independent , the probability of each element in $y$ and $z$ is (conditionally) independent as well. 
We can therefore code all variables in a latent tensor in parallel. 
This was implemented by partitioning and indexing the bitstream, providing entry points for parallel decoding~\cite{Said:15:cdo}.
Figure~\ref{fig:pec} shows an illustration of this process. 
Concurrent encoders save data to temporary memory buffers, and after frame encoding is completed, those bitstreams are concatenated to create the combined bitstream for that frame. Note that this bitstream must include a header indicating starting positions for each independent part, which are used as entry points for parallel decoding.

The coding method is an implementation of static (instead of adaptive) arithmetic coding using 32-bit registers, with 16-bit values for probabilities and range, and byte-based renormalization~\cite{Said:03:acc}. 
Since it uses only simple arithmetic and logic operations, it can be executed with multiple threads, without specialized hardware.

The integer cumulative distribution arrays needed for arithmetic coding~\cite{Said:03:acc} are pre-computed, according to the quantized entropy coding parameter defined by eq.~(\ref{eq:ECQuant}), assuming that the latent variables to be encoded have normal probability distributions, and are converted to integers using unit-step uniform quantization (i.e., simple rounding).
These arrays need to be stored at the transmitter and receiver side.

Another important factor for entropy coding on a mobile device is therefore the amount of memory needed for storing code tables. 
Throughput is maximized when coding tables stay loaded on small caches with the fastest memory. 
Fortunately, for NN-based codecs these tables are read-only (conventional codecs use adaptive coding requiring reading and writing), and the techniques and analysis developed for low-precision quantization in~\cite{Said:22:ppq} could also be employed to further reduce table memory to a few kilobytes.

\begin{figure*}
\includegraphics[width=0.745\textwidth]{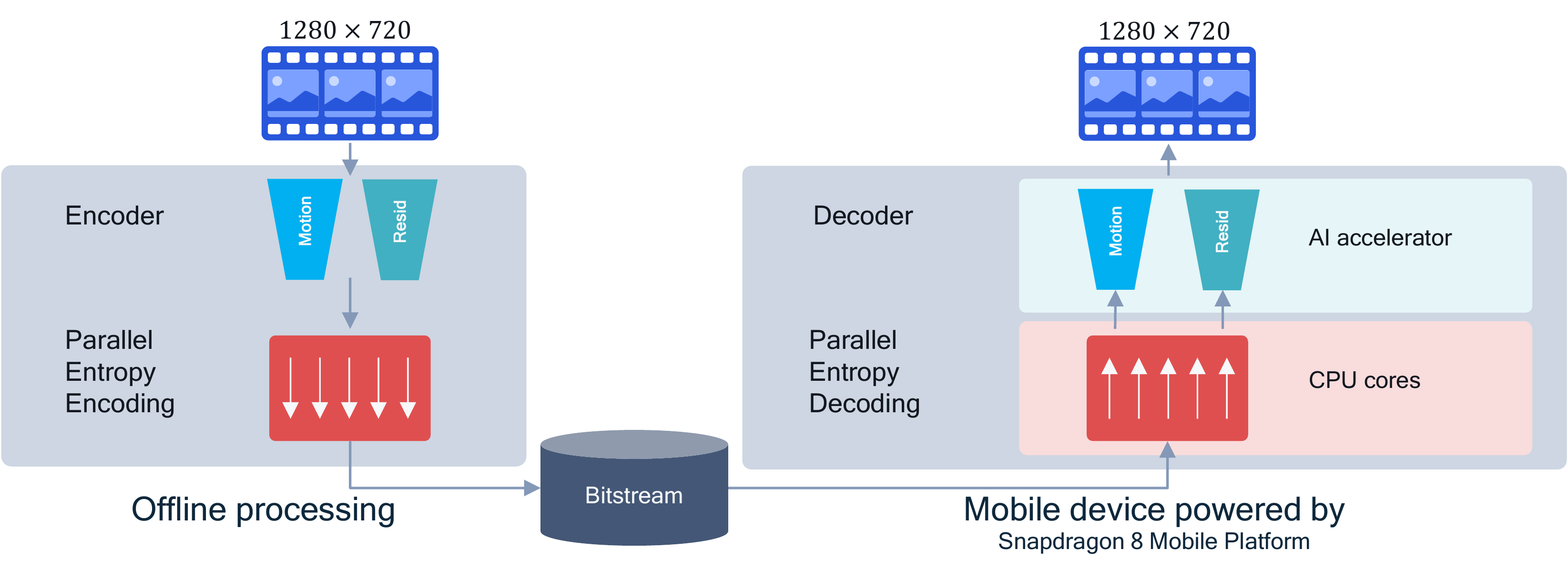}
\centering
\vspace{-0.1cm}
\caption{Configuration of our demo, showing sender (left) and receiver (right).~\label{fig:demo-config}}
\end{figure*}

\section{Experiments}

\begin{figure}
\includegraphics[width=3.25in]{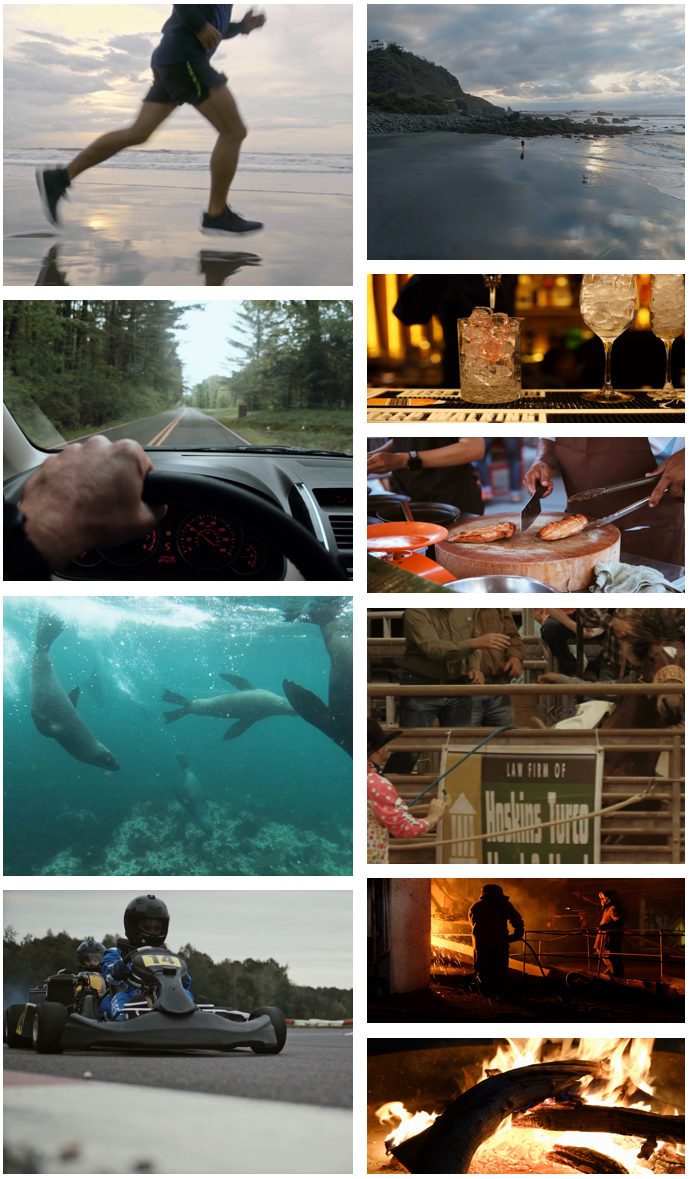}
\centering
\caption{Off-the-shelf videos collected from www.pexels.com ~\label{fig:samples}. The video set contains video of various categories of content, lighting, motion.}
\end{figure}

The goal of our experiments is to assess the ability of our neural codec to run in real-time on a mobile device while outputting high visual quality videos. 
To showcase our codec, we collected 13 off-the-shelf videos of various categories at 4k resolution from \textit{www.pexels.com}. 
Figure~\ref{fig:samples} shows snapshots of the sample videos.
These videos were then downsampled to a HD 720p resolution to reduce artifacts from compression.
As the codec and hyper-codec use six convolutions with stride two, the input resolution must be a multiple of 64. 
We therefore pad the input to $768 \times 1280$ before encoding, and the reconstructions are cropped back to $720 \times 1280$ before they are shown.

Following the training scheme for the existing Scale-Space Flow video codec~\cite{Agustsson2020-ro}, our MobileCodec was first trained on $256\times 256$ crops on the Vimeo-90k dataset~\cite{Xue_2019} for one million iterations.
The resulting floating-point model was subsequently finetuned using quantization-aware training for another 100 thousands iterations, before quantizing to 8 bits. 
The end result is an efficient 8-bit quantized neural video codec.

\subsection{Real-time HD video decoding}
We evaluated the performance of MobileCodec on the Pexels videos in a real-time decoding configuration. 
Figure~\ref{fig:demo-config} shows the setup of our experiments. 
Specifically, the input videos were encoded offline, stored as a bitstream, and then sent to a receiver, which is a mobile phone powered by a Snapdragon 8 chip. 
The receiver then decodes the bitstream into video frames at real-time speed. 
Although our demo was configured as an offline setup, the videos were encoded in low-delay mode, which could also enable live streaming in future work.

\begin{figure*}
\includegraphics[width=0.98\textwidth]{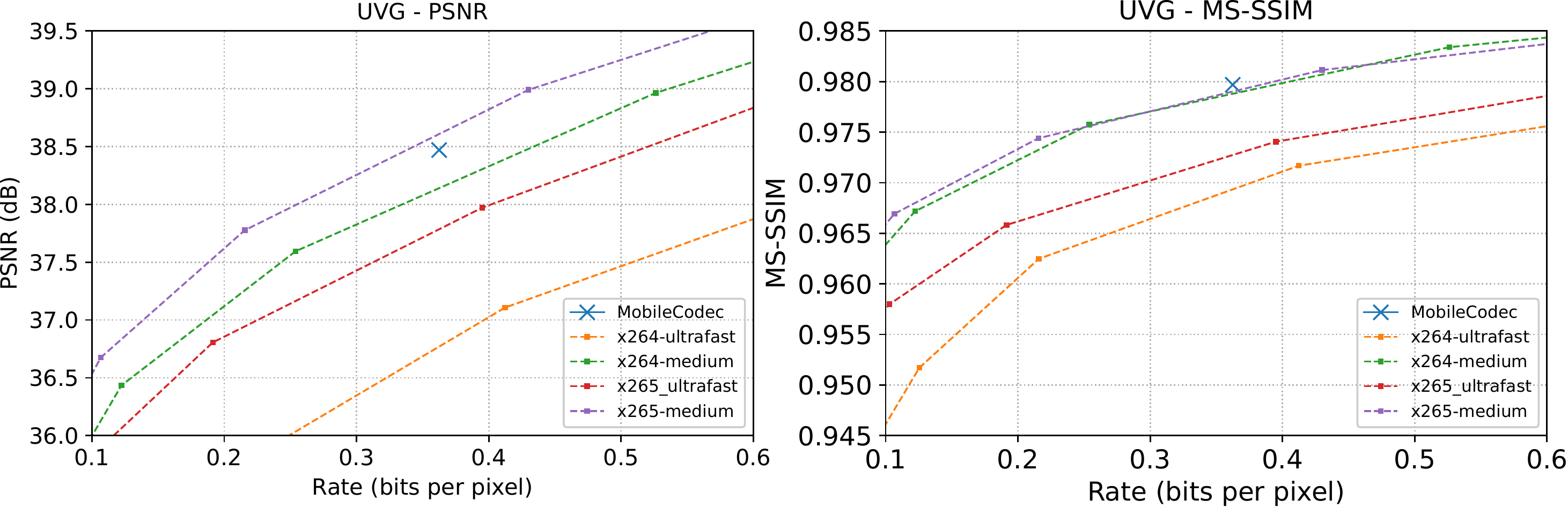}
\centering
\caption{Performance of MobileCodec on the video compression dataset UVG.~\cite{Mercat2020-dt}~\label{fig:eval-uvg}}
\end{figure*}

In Table~\ref{tab:demo-videos}, we show the quantitative measurement of our MobileCodec on each demo video. The results show that our Mobile codec can consistently run at faster than real-time speeds while obtaining satisfactory visual quality results. Specifically, our MobileCodec took 32ms to decode each frame, which is equivalent to 31 frames per second (FPS), while obtaining an average PSNR of 40.18dB.

\subsection{Quantitative Evaluation}
To further access the quality of our MobileCodec we also test its performance on the video compression dataset UVG~\cite{Mercat2020-dt}, which is widely used for benchmarking video codecs. 
To the best of our knowledge, our MobileCodec is the first neural inter-frame video codec that can run real-time decoding on a mobile device. 
Therefore, we do not compare to neural video codecs here, but do compare to fast implementations of H.264 and H.265, both among the most popular video codecs used in both regular and mobile platforms. 
Figure~\ref{fig:eval-uvg} shows the performance of our MobileCodec in comparison with the commonly used conventional codecs H264 and H265 in ffmpeg ultrafast and medium preset, tuned for fast encoding (zerolatency mode). 
This indicates that MobileCodec can obtain competitive results in comparison to the ffmpeg baselines.
Of course, stronger video codecs already exist today, both in the standards setting and in the neural codec literature. 
We believe that when these are implemented on mobile phones, we can include other codecs that are able to perform realtime decoding.
As hardware and neural inference ecosystem for mobile continues to improve, it will become increasingly easy to run neural codecs on mobile devices.

\subsection{Computational complexity}

We report the complexity of the used components in terms of network parameters and Multiply-and-Accumulate operations (MACs) in Table \ref{tab:compute}.
We compute MAC count on an artificial sequence of 2 timesteps, and normalize i    t with respect to number of input pixels to facilitate comparisons across input resolutions.
On the side of the sender, all modules need to be run in their entirety.   
On the side of the receiver, we only have to run the decoder and hyper-decoder of each component, as well as the analysis transform that processes the previous frame for motion compensation purposes.
This means that the MAC count on the receiver end is up to two times lower than on the sender side.

Note that the I-frame net only has to be run once per Group of Pictures, after which only the P-frame components are used.
This means that the total cost of decoding one P-frame, which involves running the motion receiver and residual receiver, is the sum of the P-frame components only: 257.1 KMACs per pixel.
We emphasize that the operators used in MobileCodec are highly optimized in neural accelerators, and that inference is highly parallelizable as well.

\begin{table}[t!]
\begin{tabular}{c c c c c c}
\toprule
\textbf{Video}   & \textbf{MSSSIM} & \textbf{PSNR}  & \textbf{BPP}  & \textbf{Frames} & \textbf{\begin{tabular}[c]{@{}c@{}}Time\\ (ms)\end{tabular}} \\ \midrule
Sports           & 0.9886          & 42.31          & 0.18          & 300             & 30                                                           \\
Animal           & 0.9891          & 39.55          & 0.35          & 265             & 33                                                           \\ 
Ocean            & 0.9842          & 40.21          & 0.35          & 300             & 33                                                           \\ 
Driving          & 0.9907          & 41.4           & 0.20          & 173             & 30                                                           \\ 
Driving          & 0.9904          & 40.56          & 0.31          & 274             & 33                                                           \\ 
Sports           & 0.9905          & 41.78          & 0.22          & 189             & 31                                                           \\ 
Horsebackriding  & 0.9921          & 41.55          & 0.28          & 300             & 31                                                           \\ 
Barbequing       & 0.9872          & 38.08          & 0.36          & 300             & 33                                                           \\ 
Welding          & 0.9859          & 37.56          & 0.48          & 300             & 33                                                           \\ 
Bartending       & 0.9901          & 37.89          & 0.37          & 300             & 32                                                           \\ 
Food\_Restaurant & 0.9915          & 39.16          & 0.35          & 300             & 33                                                           \\ 
Cutting          & 0.9910          & 42.62          & 0.13          & 300             & 29                                                           \\ 
Food\_Restaurant & 0.9857          & 39.67          & 0.28          & 300             & 33                                                           \\ \midrule
\textbf{Average} & \textbf{0.9890} & \textbf{40.18} & \textbf{0.30} & \textbf{277}    & \textbf{32}                                                  \\ 
\bottomrule
\end{tabular}
\caption{Performance and decoding time of MobileCodec on off-the-shelf HD videos from Pexels.~\label{tab:demo-videos}}
\end{table}

\begin{table}[t!]
    \centering
    \begin{tabular}{l r r r r}
        \toprule
        Module &  \multicolumn{2}{l}{Parameters} & \multicolumn{2}{l}{KMACs / pixel} \\
    \midrule
         I-frame net &   6.678M &    26.8\% &   211.6 &      37.0\% \\
          Motion net &  11.630M &    46.7\% &   175.6 &      30.8\% \\
        Residual net &   6.573M &    26.5\% &   183.6 &      32.2\% \\
               Total &  24.881M &   100.0\% &   570.8 &     100.0\% \\
    \midrule
    I-frame receiver &   2.857M &    24.7\% &   130.9 &      33.6\% \\
     Motion receiver &   5.980M &    51.6\% &   156.2 &      40.0\% \\
   Residual receiver &   2.752M &    23.7\% &   102.9 &      26.4\% \\
      Total receiver &  11.589M &   100.0\% &   390.1 &     100.0\% \\
    \bottomrule
    \end{tabular}
    \caption{Parameter count and MACs of MobileCodec for the sender (top) and receiver side (bottom).}
    \label{tab:compute}
\end{table}

\section{Conclusion}
This paper presents MobileCodec, an efficient neural video codec designed for deployment on a mobile platform.
Specifically, we design an inter-frame video compression architecture that performs motion compensation and transmission jointly and in feature space. 
We describe our approach to prepare this video codec for deployment, which includes a quantization-aware training step to quantize the model to 8-bit fixed point, and a parallel entropy coding algorithm that efficiently encodes neural network outputs to bitstream. 
The result is the first public demo of real-time ($\geq30$ FPS) decoding of HD 720p videos using a neural inter-frame video codec on a mobile device.

\bibliographystyle{ACM-Reference-Format}
\bibliography{sample-base}

\end{document}